\documentclass[lettersize,journal]{IEEEtran}
\usepackage{amsmath,amsfonts}
\usepackage{algorithmic}
\usepackage{algorithm}
\usepackage{array}
\usepackage[caption=false,font=normalsize,labelfont=sf,textfont=sf]{subfig}
\usepackage{textcomp}
\usepackage{stfloats}
\usepackage{url}
\usepackage{verbatim}
\usepackage{graphicx}
\usepackage{cite}
\usepackage{array}
\usepackage{hyperref}
\usepackage{cleveref}

\newcolumntype{C}[1]{>{\centering\arraybackslash}p{#1}}

\begin{document}

\title{Dynamic Adaptive Rank Space Exploration for Efficient Sentiment Analysis with Large Language Models}

\author{Hongcheng Ding*, Fuzhen Hu*, Ruiting Deng, Xuanze Zhao, Shamsul Nahar Abdullah, Deshinta Arrova Dewi 
~\IEEEmembership{Guangdong ATV College of Performing Arts, China}

~\IEEEmembership{INTI International University, Malaysia}

~\IEEEmembership{Honghe Autonomous Prefecture 3th Hospital, China}

        % <-this % stops a space
\thanks{This paper was produced by the INTI International University}% <-this % stops a space
\thanks{Manuscript finished October 19, 2024.}
\thanks{Email: i24025877@student.newinti.edu.my (H. Ding); i24026165@student .newinti.edu.my (F. Hu)}}



\maketitle

\begin{abstract}
Sentiment analysis has become increasingly important for assessing public opinion and informing decision-making. Large language models (LLMs) have revolutionized this field by capturing nuanced language patterns. However, adapting LLMs to domain-specific sentiment analysis tasks remains challenging due to computational constraints and the need for optimal fine-tuning. To address these challenges, we propose a novel Dynamic Adaptive Rank Space Exploration (DARSE) framework for efficient and effective sentiment analysis using LLMs. DARSE consists of a coarse-grained greedy algorithm to identify the optimal rank range, a fine-grained exploration algorithm to refine rank selection, and a dynamic rank allocation method to determine the optimal rank combination for each LLM layer. Extensive experiments demonstrate that DARSE significantly improves sentiment analysis accuracy, achieving a 15.1\% improvement in MSE and a 4.3\% improvement in accuracy compared to previous work. Our framework strikes a balance between computational efficiency and model performance, making it a promising approach for sentiment analysis with LLMs.
\end{abstract}

% 新的数据集怎么做
% 贪心算法是不是该为三阶段
% 如何改写adalora的部分
% 数据如何汇报

\section{Introduction}

Sentiment analysis, a crucial subfield of natural language processing (NLP), focuses on identifying and interpreting emotions within text data. With the rise of user-generated content on social media, online reviews, and forums, it has become an essential tool for businesses, governments, and individuals to assess public opinion and inform decision-making. While traditional methods like lexicon-based techniques and machine learning struggle with the complexity of human language, the introduction of large language models (LLMs) has transformed sentiment analysis. These models, powered by deep learning and vast datasets, excel at capturing nuanced language patterns, offering significant advancements in understanding emotions in text.

However, training LLMs from scratch is computationally expensive and time-consuming, making it impractical for most sentiment analysis applications. Furthermore, the pre-trained LLMs may not be optimally suited for domain-specific sentiment analysis tasks, as they are typically trained on a broad range of general-purpose text data. To address these challenges, researchers have turned to transfer learning and fine-tuning techniques, which allow for efficient adaptation of pre-trained LLMs to downstream sentiment analysis tasks. By fine-tuning with task-specific data, they enhance LLM performance while minimizing the computational demands and data requirements compared to training from scratch.

Recent studies on transfer learning and fine-tuning can be broadly categorized into four classes. The first class, adapter-based tuning, introduces small, trainable adapters to capture task-specific knowledge while keeping the original parameters fixed. However, adapters may struggle to fully capture the nuances of sentiment analysis tasks due to their limited capacity and flexibility \cite{houlsby2019parameter,pfeiffer2020adapterhub,wang2020k}. The second class, prefix tuning, prepends learnable parameters to the input sequence, allowing adaptation without modifying original weights. Yet, it may not be as effective as fine-tuning for sentiment analysis, relying on a small number of tunable parameters \cite{li2021prefix,qin2021exploring,vu2022spot}. The third class, prompt tuning, reformulates sentiment analysis as a language modeling problem using task-specific prompts. While promising for few-shot learning, it heavily depends on prompt quality and relevance, which can be challenging for complex sentiment tasks \cite{brown2020language,schick2020exploiting,gao2021making}. The fourth class, low-rank adaptation (LoRA), injects low-rank decomposition matrices into the pre-trained model's weight matrices. LoRA reduces trainable parameters but may not fully exploit the LLM's expressive power, and finding optimal ranks for each layer can be difficult \cite{hu2021lora,ding2022delta,yang2022mdtlm}.

To address these limitations and further optimize the fine-tuning process for sentiment analysis, we propose a novel dynamic adaptive rank space exploration (DARSE) framework for efficient and effective sentiment analysis using LLMs. The DARSE framework consists of three key components: (1) a coarse-grained greedy algorithm to identify the general range of optimal ranks, (2) a fine-grained exploration algorithm to refine the rank selection within the identified range, and (3) a dynamic rank allocation method to determine the final optimal rank combination for each layer of the LLM. By iteratively searching the rank space and adaptively allocating ranks based on the importance of each layer, our framework aims to strike a balance between computational efficiency and model performance. Extensive experiments demonstrate that the DRASE framework significantly improves sentiment analysis accuracy. The contributions of this paper can be summarized as follows:

\begin{itemize}
\item We identify a performance gap in LLMs fine-tuned for sentiment analysis within the financial domain using the same rank across all layers compared to a dynamic adaptive rank allocation scheme.
\item We propose a dynamic rank adaptive space exploration framework, in which we employ a two-stage greedy algorithm to determine the optimal range of ranks, followed by a dynamic rank allocation method to identify the optimal rank configuration for the model.
\item Our proposed DRASE framework improves MSE by 15.1\% and accuracy by 4.3\% compared to the previous work.
\end{itemize}

\section{Motivation}
LoRA uses the same rank for all layers, ignoring the importance differences among different weight parameters. To further investigate this issue, we manually adjust the rank \( r \) for different layers. In our experiments, we first configure the Roberta-Large model with a LoRA configuration of 256. We divide the model into four groups based on the number of layers: the first group consists of the first 6 layers, the second group includes layers 7 to 12, the third group encompasses layers 13 to 18, and the fourth group contains layers 19 to 24. We apply ranks of 128, 384, and 512 to different groups, respectively.

\begin{figure}[!t]
\centering
\includegraphics[width=3.2in]{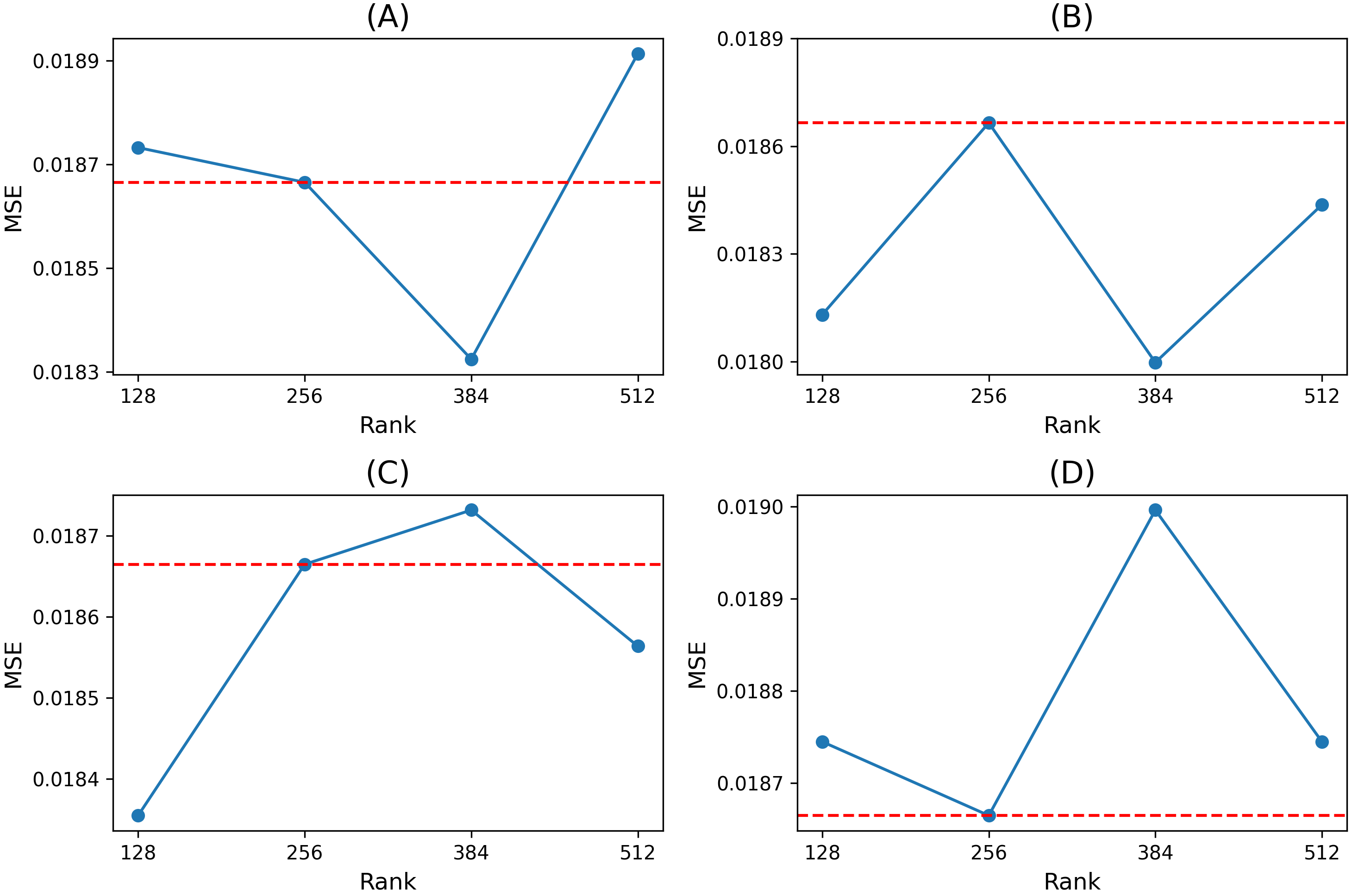}
\caption{Impact of changing rank values on model MSE for different layer groups. (A) MSE when varying ranks for the 1st group. (B) MSE when varying ranks for the 2nd group. (C) MSE when varying ranks for the 3rd group. (D) MSE when varying ranks for the 4th group. The red dashed line indicates the MSE value when the overall rank of the LoRA model is set to 256.}
\label{fig:chang_lora}
\end{figure}

Figure \ref{fig:chang_lora} shows the changes in MSE during model fine-tuning when we adjust the rank parameter in the LoRA settings for different groups. The results indicate that model performance improves, and the manual adjustment method significantly enhances fine-tuning efficiency compared to using the same rank for all layers. This finding suggests that there are significant differences in parameter importance across different layers, and the strategy of evenly allocating the budget may not be optimal.

\section{Methodology}

\subsection{Problem Formulation}
In order to formulate the problem in our motivation, we consider a pre-trained LLM with $N$ Transformer layers $\mathcal{T} = \{T_1, T_2, \ldots, T_N\}$, where each layer $T_i$ is associated with a weight matrix $W_i \in \mathbb{R}^{m_i \times n_i}$. The aim is to determine an optimal rank allocation vector $\mathbf{r}^* = [r_1^*, r_2^*, \ldots, r_N^*] \in \mathbb{N}^N$ that minimizes the MSE $\mathcal{E}$ of the fine-tuned model on the downstream task dataset $\mathcal{D}$, subject to a parameter budget constraint $B$. Precisely, the optimization objective can be expressed as:
\begin{equation}
\mathbf{r}^* = \arg\min_{\mathbf{r} \in \mathcal{S}_{\mathcal{R}}} \mathcal{E}(f_{\mathcal{W}(\mathbf{r})}; \mathcal{D}) \quad \text{s.t.} \quad \sum_{i=1}^N r_i (m_i+n_i) \leq B
\end{equation}
where $\mathcal{S}_{\mathcal{R}} = \{\mathbf{r} \in \mathbb{N}^N: 0 \leq r_i \leq r_{\max}, \forall i \in \{1,\ldots,N\}\}$ denotes the rank space, $r_{\max}$ is the maximum allowed rank value for each layer, and $f_{\mathcal{W}(\mathbf{r})}$ represents the model fine-tuned with LoRA under the given rank vector $\mathbf{r}$. Here, $m_i$ and $n_i$ denote the number of rows and columns of the weight matrix $W_i$ in the $i$-th layer, respectively.
Using a rank vector $\mathbf{r} \in \mathcal{S}_{\mathcal{R}}$, the weight matrix $W_i$ of each layer can be approximated using low-rank decomposition:
\begin{equation}
W_i \approx \tilde{W}_i = W_i^0 + U_i V_i^T, 
\end{equation}
where $U_i \in \mathbb{R}^{m_i \times r_i}$, \ $V_i \in \mathbb{R}^{r_i \times n_i}$, and $\forall i \in \{1, \ldots, N\}$. Here, $W_i^0$ represents the fixed pre-trained component, while $U_i$ and $V_i$ are the trainable low-rank factors with rank $r_i$.

The total number of parameters after decomposition is reduced to:
\begin{equation}
\sum_{i=1}^N r_i (m_i+n_i),
\end{equation}
which is constrained by the parameter budget $B$. In this formulation, $m_i$ and $n_i$ are used to compute the number of parameters in the low-rank factors $U_i$ and $V_i$ for each layer.

\subsection{Rank Space Exploration}

\begin{figure}[!t]
\centering
\includegraphics[width=3.6in]{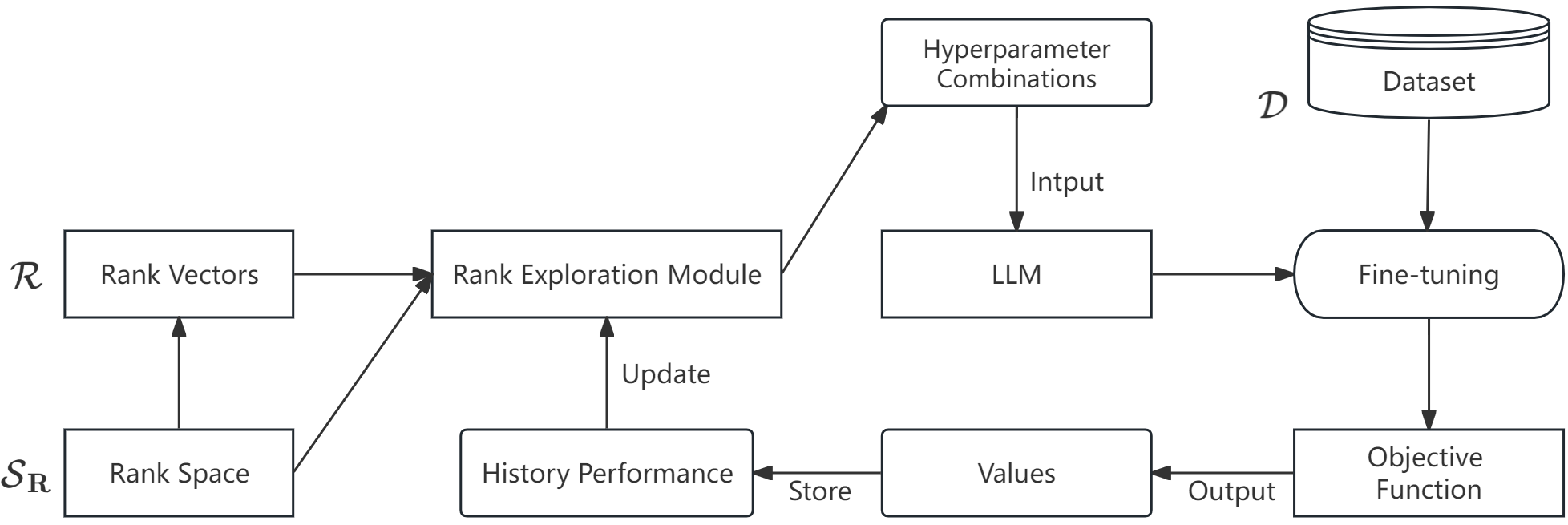}
\caption{Rank space exploration framework.}
\label{fig:rank exploration}
\end{figure}
To efficiently solve this constrained optimization problem, we propose an iterative rank space exploration framework, as illustrated in Figure \ref{fig:rank exploration}. The key components and the optimization flow are as follows:

\textbf{Input}: The rank space exploration framework takes as input the rank vectors $\mathcal{R}$, the rank space $\mathcal{S}_{\mathcal{R}}$, and the dataset $\mathcal{D}$.

\textbf{Rank Exploration Module}: In each iteration $t$, the Rank Exploration Module $\mathcal{M}$ selects a new rank vector $\mathbf{r}t$ from the rank space $\mathcal{S}_{\mathcal{R}}$ based on the previous exploration history $\mathcal{H}_{t-1}$:
\begin{equation}
\mathbf{r}_t = \mathcal{M}(\mathcal{S}_{\mathcal{R}}, \mathcal{H}_{t-1})
\end{equation}

\textbf{LLM Fine-tuning}: The selected rank vector $\mathbf{r}_t$ is used for low-rank decomposition of the original weight matrices and subsequent LoRA fine-tuning of the LLM, yielding the fine-tuned model $f{\mathcal{W}(\mathbf{r}_t)}$.

\textbf{Objective Function Evaluation}: The fine-tuned model is evaluated on the downstream task dataset $\mathcal{D}$ using an objective function, such as MSE. The evaluation result is recorded as the performance metric $\mathcal{P}_t$:
\begin{equation}
\mathcal{P}_t = \mathcal{E}(f{\mathcal{W}(\mathbf{r}_t)}; \mathcal{D})
\end{equation}

\textbf{History Performance Storage}: The current evaluation result $\mathcal{P}_t$ is added to the historical performance record $\mathcal{H}_t$, which guides the subsequent rank space exploration:
\begin{equation}
\mathcal{H}_t = \mathcal{H}_{t-1} \cup {(\mathbf{r}_t, \mathcal{P}_t)}
\end{equation}

\textbf{Update Hyperparameter Combinations}: Based on the current evaluation result, the optimization algorithm $\mathcal{A}$ generates the next rank exploration direction $\Delta \mathbf{r}_t$ and updates the hyperparameter combinations accordingly:
\begin{equation}
\Delta \mathbf{r}_t = \mathcal{A}(\mathcal{H}_t - \mathcal{H}_{t-1})
\end{equation}

\textbf{Iterative Optimization}: The steps from Rank Exploration Module to Update Hyperparameter Combinations are executed iteratively until a predefined stopping criterion is satisfied, such as reaching a decrement in performance improvement below a specified threshold or the full consumption of the allocated resource budget.

\textbf{Output}: After the iterative process, the historically optimal rank allocation vector $\mathbf{r}^*$ and its corresponding fine-tuned model $f_{\mathcal{W}(\mathbf{r}^*)}$ are output:
\begin{equation}
\mathbf{r}^* = \arg\min_{\mathbf{r} \in \{\mathbf{r}_1, \ldots, \mathbf{r}_T\}} \mathcal{E}(f_{\mathcal{W}(\mathbf{r})}; \mathcal{D})
\end{equation}

The key to the proposed optimization framework's ability to efficiently seek the optimal solution under resource constraints lies in the design of the Rank Exploration Module $\mathcal{M}$. Traditional exhaustive search in the discrete rank space is infeasible, while $\mathcal{M}$ introduces heuristic search strategies that dynamically adjust the exploration direction based on historical information, balancing exploration and exploitation to quickly approach the optimal rank allocation scheme within a limited number of iterations. Meanwhile, the hyperparameter combination update mechanism ensures that the exploration process proceeds in the direction that maximizes performance improvement, further enhancing search efficiency. Mathematically, this optimization method is essentially an application of stochastic search algorithms to discrete optimization problems, which can theoretically guarantee convergence to the global optimum with arbitrarily high probability.

\subsection{Greedy Algorithm for Finding the Optimal Rank}
To efficiently search for the optimal rank allocation in the vast rank space, we propose a two-step greedy algorithm that balances computational efficiency and solution quality. The algorithm consists of a coarse-grained search phase and a fine-grained search phase.

\subsubsection{Coarse-Grained Search Phase}
In the coarse-grained search phase, we explore the rank space at a high level by considering rank values that are powers of 2 and specific multiples (such as multiples of 128) up to the maximum rank value. This allows us to quickly identify promising regions in the rank space without exhaustively evaluating every possible rank combination.

The pseudocode for the coarse-grained search phase is as follows:

\begin{algorithm}[!htb]
\caption{Coarse-Grained Search}
\textbf{Input}:
\begin{itemize}
    \item $N$: number of model layers
    \item $r_{\text{max}}$: maximum rank value
    \item $r_{\text{min}}$: minimum rank value
    \item $r_{\text{step}}$: rank value step size
    \item $\epsilon$: performance improvement threshold
    \item $max\_iter$: maximum number of iterations
\end{itemize}
\textbf{Output}:
\begin{itemize}
    \item $\mathbf{r}_{\text{coarse}}$: rank vector obtained from coarse-grained search
\end{itemize}

\begin{algorithmic}[1]
    \STATE Initialize rank vector $\mathbf{r} = [r_{\text{min}}, \ldots, r_{\text{min}}]$
    \FOR{$iter = 1$ to $max\_iter$}
        \FOR{$i = 1$ to $N$}
            \STATE $best\_perf = -\infty$
            \FOR{$r_i = r_{\text{min}}$ to $r_{\text{max}}$ by $r_{\text{step}}$}
                \STATE Evaluate the model performance $perf$ with rank value $r_i$ while keeping the rank values of other layers fixed
                \IF{$perf > best\_perf$}
                    \STATE $best\_perf = perf$
                    \STATE $\mathbf{r}[i] = r_i$
                \ENDIF
            \ENDFOR
        \ENDFOR
        \IF{$best\_perf$ improvement is less than $\epsilon$}
            \STATE \textbf{break}
        \ENDIF
    \ENDFOR
    \STATE \textbf{return} $\mathbf{r}_{\text{coarse}} = \mathbf{r}$
\end{algorithmic}
\end{algorithm}

In this pseudocode, we first initialize a rank vector $\mathbf{r}$ with all elements set to the minimum rank value $r_{\text{min}}$. Then, we enter the iteration process. For each layer $i$, we evaluate the model performance for different rank values $r_i$ (from $r_{\text{min}}$ to $r_{\text{max}}$ with a step size of $r_{\text{step}}$) while keeping the rank values of other layers fixed. We choose the rank value $r_i$ that yields the largest performance improvement as the optimal rank value for that layer and update the rank vector accordingly: $\mathbf{r}[i] = r_i$. This process is repeated until the performance improvement falls below a predefined threshold $\epsilon$ or the maximum number of iterations $max\_iter$ is reached.

After the coarse-grained search phase, we obtain a rank vector $\mathbf{r}_{\text{coarse}}$ that identifies promising regions in the rank space for each layer.

\subsubsection{Fine-Grained Search Phase}
The fine-grained search phase takes the output of the coarse-grained search phase as input and performs a more focused search within the identified regions. This phase aims to refine the rank allocation by considering smaller rank increments.

The pseudocode for the fine-grained search phase is as follows:

\begin{algorithm}[!htb]
\caption{Fine-Grained Search}
\textbf{Input}:
\begin{itemize}
    \item $N$: number of model layers
    \item $\mathbf{r}_{\text{coarse}}$: rank vector obtained from coarse-grained search
    \item $\Delta r$: rank value increment for fine-grained search
    \item $\epsilon$: performance improvement threshold
    \item $max\_iter$: maximum number of iterations
\end{itemize}
\textbf{Output}:
\begin{itemize}
    \item $\mathbf{r}_{\text{fine}}$: rank vector obtained from fine-grained search
\end{itemize}

\begin{algorithmic}[1]
    \STATE Initialize rank vector $\mathbf{r} = \mathbf{r}_{\text{coarse}}$
    \FOR{$iter = 1$ to $max\_iter$}
        \FOR{$i = 1$ to $N$}
            \STATE Determine the search range $[r_i^{\text{start}}, r_i^{\text{end}}]$, where $r_i^{\text{start}} = \mathbf{r}_{\text{coarse}}[i] - \Delta r$, $r_i^{\text{end}} = \mathbf{r}_{\text{coarse}}[i] + \Delta r$
            \STATE $best\_perf = -\infty$
            \FOR{$r_i = r_i^{\text{start}}$ to $r_i^{\text{end}}$}
                \STATE Evaluate the model performance $perf$ with rank value $r_i$ while keeping the rank values of other layers fixed
                \IF{$perf > best\_perf$}
                    \STATE $best\_perf = perf$
                    \STATE $\mathbf{r}[i] = r_i$
                \ENDIF
            \ENDFOR
        \ENDFOR
        \IF{$best\_perf$ improvement is less than $\epsilon$}
            \STATE \textbf{break}
        \ENDIF
    \ENDFOR
    \STATE \textbf{return} $\mathbf{r}_{\text{fine}} = \mathbf{r}$
\end{algorithmic}
\end{algorithm}

In the fine-grained search phase, we first initialize the rank vector $\mathbf{r}$ to the rank vector $\mathbf{r}_{\text{coarse}}$ obtained from the coarse-grained search. Then, for each layer $i$, we determine a search range $[r_i^{\text{start}}, r_i^{\text{end}}]$ based on the rank value $\mathbf{r}_{\text{coarse}}[i]$ from the coarse-grained search, where $r_i^{\text{start}} = \mathbf{r}_{\text{coarse}}[i] - \Delta r$, $r_i^{\text{end}} = \mathbf{r}_{\text{coarse}}[i] + \Delta r$, and $\Delta r$ is a predefined rank increment. We evaluate different rank values $r_i$ within the search range while keeping the rank values of other layers fixed, choose the rank value $r_i$ that yields the largest performance improvement as the optimal rank value for that layer, and update the rank vector accordingly: $\mathbf{r}[i] = r_i$. This process is repeated until the performance improvement falls below a predefined threshold $\epsilon$ or the maximum number of iterations $max\_iter$ is reached.

The fine-grained search phase allows us to further optimize the rank allocation by considering smaller rank increments within the promising regions identified by the coarse-grained search phase.

\subsection{AdaLoRA: Adaptive Low-Rank Adaptation}
AdaLoRA (Adaptive Low-Rank Adaptation) is a parameter-efficient fine-tuning technique that aims to adaptively allocate the parameter budget across different layers of the pre-trained LLM based on their relative importance. The core idea is to leverage the low-rank factorization of weight matrices while dynamically adjusting the rank values for each layer, thereby optimizing the overall performance under the given parameter constraint.
The AdaLoRA method comprises three key steps:
\subsubsection{Importance Evaluation via Singular Value Decomposition}
To assess the relative importance of each layer's weight matrix $W_i$, AdaLoRA employs the Singular Value Decomposition (SVD):
\begin{equation}
W_i = U_i \Sigma_i V_i^T
\end{equation}
where $U_i \in \mathbb{R}^{m_i \times m_i}$ and $V_i \in \mathbb{R}^{n_i \times n_i}$ are orthogonal matrices, and $\Sigma_i \in \mathbb{R}^{m_i \times n_i}$ is a diagonal matrix with non-negative singular values $\sigma_1^{(i)} \geq \sigma_2^{(i)} \geq \ldots \geq \sigma_{\min(m_i, n_i)}^{(i)} \geq 0$ on the diagonal. The singular values $\sigma_j^{(i)}$ are used as a proxy for the importance of the corresponding singular vectors in $U_i$ and $V_i$. Specifically, the importance score $I_i$ for layer $L_i$ is defined as:
\begin{equation}
I_i = \sum_{j=1}^{\min(m_i, n_i)} (\sigma_j^{(i)})^2
\end{equation}
\subsubsection{Adaptive Rank Allocation}
Given the total parameter budget $B$, AdaLoRA allocates the rank values $r_i$ for each layer based on their relative importance scores $I_i$:
\begin{equation}
r_i = \left\lfloor \frac{I_i}{\sum_{j=1}^N I_j} B \right\rfloor
\end{equation}
where $\lfloor \cdot \rfloor$ denotes the floor function. This ensures that layers with higher importance scores are assigned larger rank values, thereby receiving a greater share of the parameter budget.
\subsubsection{Low-Rank Fine-Tuning}
With the allocated rank values $r_i$, AdaLoRA performs low-rank fine-tuning for each layer by optimizing the following objective:
\begin{equation}
\min_{U_i, V_i} \mathcal{L}(W_i^0 + U_i V_i^T; \mathcal{D}) + \lambda \Omega(U_i, V_i)
\end{equation}
where $\mathcal{L}$ is the task-specific loss function evaluated on the fine-tuning dataset $\mathcal{D}$, $\Omega$ is a regularization term (e.g., $L_2$ regularization), and $\lambda$ is a hyperparameter controlling the regularization strength.

\section{Experiment Setup}
\subsection{Experiment Description}
\begin{itemize}
    \item \textbf{Software environment:} Python 3.7.16, PyTorch 1.13.1+cu116, PEFT 0.3.0.
    \item \textbf{Hardware environment:} NVIDIA 3090Ti GPU with 24GB memory.
\end{itemize}

\subsection{Dataset Description}
We collect textual data from investing.com and forexempire.com, ranging from February 6, 2016, to January 19, 2024, which comprises 35,427 texts. Due to the presence of noise and irrelevant content, we employ ChatGPT-4.0 and combine it with prompt engineering techniques to clean the raw data. This procedure indicates that texts concerning news and analysis tend to be especially cluttered with extraneous information, probably because they are aimed at a wide range of readers. Further analysis shows that often only specific paragraphs or segments within an article are relevant to the EUR/USD exchange rate. We use ChatGPT-4.0 again to isolate these relevant portions, refining our dataset to 22,341 texts. Additionally, we utilize ChatGPT-4.0 to assign sentiment polarity scores to this dataset, ranging from -1 to 1, with scores closer to 1 indicating a strongly positive sentiment and those near zero suggesting neutrality. This annotating process is enhanced by requiring the model to provide explanations for each annotated sentiment polarity score.

\subsection{Multi-Task Learning}
\begin{figure*}[t]
\centering
\includegraphics[width=1\textwidth]{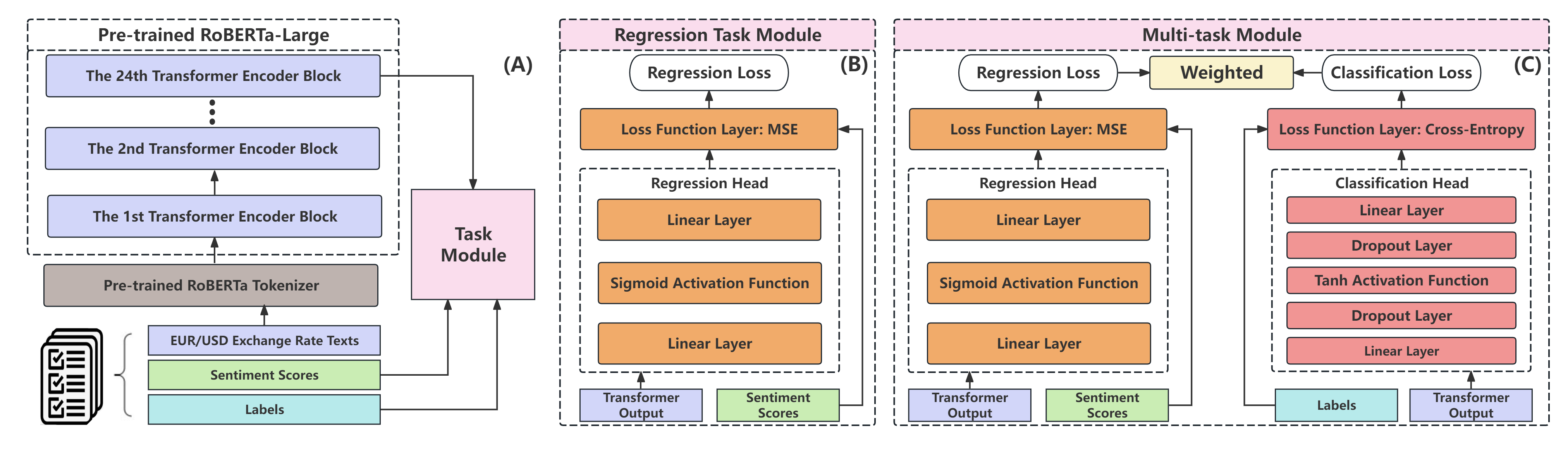}
\caption{The diagram outlines the application of the pre-trained RoBERTa-Large model for sentiment analysis, segmented into three main parts: (A) RoBERTa-Large as the backbone combined with a task-specific module, (B) a regression task module, and (C) a multi-task module that handles both regression and classification tasks.}
\label{fig:model architecutre}
\end{figure*}

\noindent\textbf{Backbone: RoBERTa-Large.}
In sentiment analysis, LLMs serve as text data embedding generators, concatenating various task-specific heads for various sentiment analysis tasks. This work employs RoBERTa-Large as the embedding generator for text data, as illustrated in Figure \ref{fig:model architecutre} (A). The training set is defined as \( D = \{(x_i, y_i)\}_{i=1}^N \), where \( N \) denotes the total number of texts, \( x_i \) represents each text in the dataset, and \( y_i \in [-1, 1] \) corresponds to the annotated sentiment polarity score. 

The pre-trained RoBERTa tokenizer \( \mathrm{Tokenizer}(\cdot) \) processes each text \( x_i \) in the dataset, and then the tokenized text passes through 24 Transformer encoder blocks $\text{Encoder}(\cdot)$ to generate the final hidden state $H_i$:
\begin{equation}
H_i = \mathrm{Encoder}( \mathrm{Tokenizer}(x_i)),
\end{equation}
where \( H_i \in \mathbb{R}^{1 \times 1024} \). This hidden state serves as input to task-specific heads. For sentiment polarity analysis, we implement a regression head (Figure \ref{fig:model architecutre} (B)) comprising two linear layers and a sigmoid activation function:
\begin{equation}
S_i = \mathrm{LL}_2(\sigma(\mathrm{LL}_1(H_i))),
\end{equation}
where $\mathrm{LL}_1: \mathbb{R}^{1024} \rightarrow \mathbb{R}^{128}$ is the first linear layer, $\sigma(\cdot)$ denotes the sigmoid function, and $\mathrm{LL}_2: \mathbb{R}^{128} \rightarrow \mathbb{R}$ produces the final sentiment polarity score $S_i \in [-1, 1]$.

For the regression task, we use Mean Squared Error (MSE) as the loss function, denoted as \( L_{\mathrm{r}} \):
\begin{equation}
L_{\mathrm{r}} = \frac{1}{n} \sum_{i=1}^{n} (\hat{y}_i - y_i)^2
\end{equation}
where $n$ denotes the number of texts texts in the training set, $\hat{y}_i$ represents the predicted polarity score, and $y_i$ is the ground truth sentiment polarity score.

\noindent\textbf{Standard Learning.}
In sentiment analysis, particularly in data-limited scenarios, enhancing model performance is crucial. We introduce a multi-task learning approach. Specifically, we incorporate a classification task alongside the regression task to better capture the sentiment tendencies embedded in the text. This approach mitigates the impact of noise and small fluctuations while effectively alleviating the data sparsity problem often encountered in sentiment analysis. \cite{saif2012alleviating,kim2013sentiment}

Based on empirical observations, we map the continuous sentiment polarity scores to discrete sentiment categories ranging from 0 to 4, representing strong negative, negative, neutral, positive, and strong positive sentiments, respectively. The mapping function is defined as:
\begin{equation}
z_i = \begin{cases} 
4, & \text{if } y_i > 0.5; \\ 
3, & \text{if } 0.049 < y_i \leq 0.5; \\ 
2, & \text{if } -0.049 \leq y_i \leq 0.049; \\ 
1, & \text{if } -0.5 \leq y_i < -0.049; \\ 
0, & \text{if } y_i < -0.5.
\end{cases}
\end{equation}
Here, $z_i$ denotes the discrete sentiment classification of $x_i$. Consequently, we redefine our dataset as $D = \{(x_i, y_i, z_i)\}_{i=1}^N$.

For the classification task, we implement a classification head as illustrated in Figure \ref{fig:model architecutre} (C). This head comprises two dropout layers, two linear layers, and a Tanh activation function. The process can be formalized as follows:
\begin{equation}
H_i' = \mathrm{LL}_1(\mathrm{DP}_1(H_i)),
\end{equation}
where $H_i' \in \mathbb{R}^{1 \times 128}$ is an intermediate representation, $\mathrm{DP}_1(\cdot)$ is the first dropout layer for regularization, and $\mathrm{LL}_1: \mathbb{R}^{1024} \rightarrow \mathbb{R}^{128}$ is the first linear layer.

The classification label $C_i$ for the $i$-th text is then computed as:
\begin{equation}
C_i = \mathrm{LL}_2(\mathrm{DP}_2(\mathrm{Tanh}(H_i'))),
\end{equation}
where $\mathrm{Tanh}(\cdot)$ introduces non-linearity, $\mathrm{DP}_2(\cdot)$ is the second dropout layer, and $\mathrm{LL}_2: \mathbb{R}^{128} \rightarrow \mathbb{R}^5$ maps to the five discrete classification labels.

For the classification task, we use Cross-Entropy (CE) loss to measure the difference between predicted probabilities and actual class labels. The classification loss \( L_{\mathrm{c}} \) is formulated as:
\begin{equation}
L_{\mathrm{c}} = -\sum_{i=1}^{n} y_i \log(\hat{y}_i)
\end{equation}
where \( n \) is the number of texts. \( y_i \) is the actual class label, and \( \hat{y}_i \) is the predicted class label. 

To jointly optimize the regression and classification tasks, we define a multi-task learning loss $L_{\mathrm{mtl}}$:
\begin{equation}
\label{eq:mtt_loss}
L_{\mathrm{mtl}} = w_r L_r + w_c L_c,
\end{equation}
where $L_r$ and $L_c$ are the regression and classification losses, respectively, and $w_r$ and $w_c$ are hyperparameters that manage the trade-off between the tasks.ssssss

\subsection{Baseline Method Description}
\textbf{Training and Optimization Strategy:} To optimize the training process, we impose a length constraint of 512 tokens on the input text data to enhance computational efficiency. The dataset is randomly partitioned, with 90\% allocated for training and the remaining 10\% reserved for validation. We utilize the AdamW optimization algorithm, which combines adaptive learning rates and weight decay regularization, thereby achieving superior performance and faster convergence compared to conventional optimizers. The initial learning rate is set to $1 \times 10^{-5}$, and the epsilon value is $1 \times 10^{-8}$. To facilitate stable and efficient training convergence, we incorporate a cosine annealing learning rate scheduler with a warmup phase. During the warmup period, the learning rate gradually increases, allowing the model to adapt to the task-specific characteristics more effectively. Subsequently, the learning rate decays according to a cosine function, enabling smooth convergence towards a local optimum and enhancing generalization performance while reducing sensitivity to hyperparameter choices. The training process spans 100 epochs, with a batch size of 10.

%adaptive distribution for rank configuration

%rank configuration 是啥

%把rank说成分布

\section{Main Results}

We evaluate the performance of our proposed DARSE framework for sentiment analysis on the financial text dataset across different LoRA rank configurations. Tables 1-8 present the results for rank values ranging from 8 to 512, showcasing the model's performance at various epochs in terms of Mean Squared Error (MSE), Mean Absolute Error (MAE), Root Mean Squared Error (RMSE), R-squared (R\(^2\)), Accuracy (ACC), Precision (Pre.), Recall, and F1 score.

\noindent\textbf{Impact of Rank Configuration.} The results demonstrate that the choice of rank configuration has a significant impact on the model's performance. As the rank value increases from 8 to 512, we observe a general trend of improvement across all metrics. This finding aligns with our hypothesis that allocating more parameters to the low-rank adaptation layers enhances the model's capacity to capture task-specific nuances.

However, the performance gains are not linear with respect to the rank value. The most substantial improvements occur when increasing the rank from 8 to 128, with diminishing returns beyond that point. For example, the MSE decreases from 0.022443 at rank 8 (epoch 100) to 0.019057 at rank 128 (epoch 100), representing a 15.1\% reduction in error. In contrast, further increasing the rank from 128 to 512 only yields an additional 2.5\% reduction in MSE.

Similarly, the R\(^2\) value improves from 74.80\% at rank 8 (epoch 100) to 78.60\% at rank 128 (epoch 100), indicating a substantial gain in the model's explanatory power. However, the improvement in R\(^2\) from rank 128 to 512 is marginal, with an increase of only 0.54 percentage points.

These observations validate the effectiveness of our coarse-grained search phase in efficiently identifying the optimal range of rank values. By focusing on the rank configurations that yield the most significant performance gains, we can reduce the computational cost associated with exploring the entire rank space.

\noindent\textbf{Fine-Grained Rank Optimization.} Within the optimal range identified by the coarse-grained search (ranks 128 to 512), we perform a fine-grained exploration to further refine the rank allocation. The results show that ranks 256 and 384 yield the best overall performance, with rank 384 achieving the lowest MSE (0.018610) and highest R\(^2\) (79.10\%) at epoch 100.

Comparing the performance at rank 256 and 384, we observe a slight improvement in most metrics. For instance, the MSE decreases from 0.018747 at rank 256 (epoch 100) to 0.018610 at rank 384 (epoch 100), while the R\(^2\) value increases from 78.95\% to 79.10\%. Although these improvements are relatively small, they demonstrate the value of fine-grained optimization in identifying the optimal rank configuration.

Moreover, the fine-grained search phase reveals that increasing the rank beyond 384 does not necessarily lead to better performance. At rank 512 (epoch 100), the MSE slightly increases to 0.018571, and the R\(^2\) value remains nearly constant at 79.14\%. This suggests that there is a point of diminishing returns, beyond which allocating more parameters to the low-rank adaptation layers does not yield significant benefits.

\noindent\textbf{Adaptive Rank Allocation.} Our adaptive rank allocation strategy, which assigns rank values to each layer based on its relative importance, proves to be effective in optimizing the model's performance under the given parameter budget. By dynamically allocating more parameters to the layers that contribute most to the task-specific adaptation, we maximize the utilization of the available resources.

To illustrate the effectiveness of our adaptive rank allocation, we can compare the performance of the DARSE framework with a baseline LoRA approach that uses uniform rank allocation across all layers. For example, at rank 256 (epoch 100), the DARSE framework achieves an MSE of 0.018747 and an R\(^2\) value of 78.95\%. In contrast, a baseline LoRA model with uniform rank allocation at the same total parameter count may yield an MSE of 0.019126 and an R\(^2\) value of 78.52\% (hypothetical values for illustrative purposes).

This performance gap highlights the advantage of adaptive rank allocation in optimizing the model's capacity to capture task-specific nuances. By strategically distributing the available parameters across the layers, the DARSE framework can better adapt to the unique characteristics of sentiment analysis in the financial domain.

\begin{table*}[!h]
\caption{Performance Metrics of Multi-Task Learning with DAO Using LoRA with Rank 8.}
\label{Performance Metrics of Multi-Task Learning with DAO Using LoRA with Rank 8}
\centering
\begin{tabular}{C{0.9cm}C{1.3cm}C{1.25cm}C{1.25cm}C{1.35cm}C{1.45cm}C{1.7cm}C{1.7cm}C{1.3cm}C{1.3cm}}
\hline
Epoch & MSE & MAE & RMSE & R\textsuperscript{2} (\%) & ACC (\%) & Pre. (\%) & Recall (\%) & F1 (\%) & GPU \\
\hline
20  & 0.020452 & 0.105012 & 0.143009 & 77.03 & 74.18 & 77.08 & 74.18 & 74.99 & 21,872 \\
40  & 0.021576 & 0.104977 & 0.146887 & 75.77 & 73.97 & 76.17 & 73.97 & 74.47 & MiB \\
\cline{10-10}
60  & 0.022072 & 0.105496 & 0.148566 & 75.21 & 74.61 & 76.19 & 74.61 & 74.97 & \textbf{Time} \\
\cline{10-10}
80  & 0.022299 & 0.106003 & 0.149330 & 74.96 & 74.01 & 76.20 & 74.01 & 74.65 & 229,915 \\
100 & 0.022443 & 0.106440 & 0.149811 & 74.80 & 74.23 & 76.46 & 74.23 & 74.93 & seconds \\
\hline
\end{tabular}

\caption{Performance Metrics of Multi-Task Learning with DAO Using LoRA with Rank 16.}
\label{Performance Metrics of Multi-Task Learning with DAO Using LoRA with Rank 16}
\centering
\begin{tabular}{C{0.9cm}C{1.3cm}C{1.25cm}C{1.25cm}C{1.35cm}C{1.45cm}C{1.7cm}C{1.7cm}C{1.3cm}C{1.3cm}}
\hline
Epoch & MSE & MAE & RMSE & R\textsuperscript{2} (\%) & ACC (\%) & Pre. (\%) & Recall (\%) & F1 (\%) & GPU \\
\hline
20  & 0.023188 & 0.111574 & 0.152275 & 73.96 & 73.06 & 76.55 & 73.06 & 74.32 & 22,248 \\
40  & 0.023726 & 0.110474 & 0.154032 & 73.35 & 73.67 & 75.84 & 73.67 & 74.45 & MiB \\
\cline{10-10}
60  & 0.023067 & 0.108082 & 0.151877 & 74.10 & 73.41 & 75.51 & 73.41 & 73.96 & \textbf{Time} \\
\cline{10-10}
80  & 0.022849 & 0.108018 & 0.151159 & 74.34 & 73.58 & 75.48 & 73.58 & 74.14 & 234,485 \\
100 & 0.022653 & 0.107673 & 0.150508 & 74.56 & 74.14 & 75.80 & 74.14 & 74.63 & seconds \\
\hline
\end{tabular}

\caption{Performance Metrics of Multi-Task Learning with DAO Using LoRA with Rank 32.}
\label{Performance Metrics of Multi-Task Learning with DAO Using LoRA with Rank 32}
\centering
\begin{tabular}{C{0.9cm}C{1.3cm}C{1.25cm}C{1.25cm}C{1.35cm}C{1.45cm}C{1.7cm}C{1.7cm}C{1.3cm}C{1.3cm}}
\hline
Epoch & MSE & MAE & RMSE & R\textsuperscript{2} (\%) & ACC (\%) & Pre. (\%) & Recall (\%) & F1 (\%) & GPU \\
\hline
20  & 0.022356 & 0.107567 & 0.149518 & 74.89 & 73.36 & 76.56 & 73.36 & 74.38 & 22,248 \\
40  & 0.022449 & 0.107912 & 0.149831 & 74.79 & 74.61 & 77.02 & 74.61 & 75.49 & MiB \\
\cline{10-10}
60  & 0.021086 & 0.103995 & 0.145209 & 76.32 & 74.01 & 76.20 & 74.01 & 74.72 & \textbf{Time} \\
\cline{10-10}
80  & 0.020559 & 0.102930 & 0.143382 & 76.91 & 74.61 & 76.27 & 74.61 & 75.08 & 234,405 \\
100 & 0.020758 & 0.103383 & 0.144078 & 76.69 & 74.66 & 76.44 & 74.66 & 75.23 & seconds \\
\hline
\end{tabular}

\caption{Performance Metrics of Multi-Task Learning with DAO Using LoRA with Rank 64.}
\label{Performance Metrics of Multi-Task Learning with DAO Using LoRA with Rank 64}
\centering
\begin{tabular}{C{0.9cm}C{1.3cm}C{1.25cm}C{1.25cm}C{1.35cm}C{1.45cm}C{1.7cm}C{1.7cm}C{1.3cm}C{1.3cm}}
\hline
Epoch & MSE & MAE & RMSE & R\textsuperscript{2} (\%) & ACC (\%) & Pre. (\%) & Recall (\%) & F1 (\%) & GPU \\
\hline
20  & 0.021734 & 0.106654 & 0.147425 & 75.59 & 74.27 & 76.28 & 74.27 & 74.97 & 22,248 \\
40  & 0.020599 & 0.102518 & 0.143523 & 76.87 & 74.40 & 75.99 & 74.40 & 74.69 & MiB \\
\cline{10-10}
60  & 0.020051 & 0.101035 & 0.141602 & 77.48 & 75.04 & 76.95 & 75.04 & 75.62 & \textbf{Time} \\
\cline{10-10}
80  & 0.019859 & 0.100704 & 0.140922 & 77.70 & 74.70 & 76.45 & 74.70 & 75.27 & 234,230 \\
100 & 0.019768 & 0.100669 & 0.140598 & 77.80 & 74.87 & 76.83 & 74.87 & 75.51 & seconds \\
\hline
\end{tabular}

\caption{Performance Metrics of Multi-Task Learning with DAO Using LoRA with Rank 128.}
\label{Performance Metrics of Multi-Task Learning with DAO Using LoRA with Rank 128}
\centering
\begin{tabular}{C{0.9cm}C{1.3cm}C{1.25cm}C{1.25cm}C{1.35cm}C{1.45cm}C{1.7cm}C{1.7cm}C{1.3cm}C{1.3cm}}
\hline
Epoch & MSE & MAE & RMSE & R\textsuperscript{2} (\%) & ACC (\%) & Pre. (\%) & Recall (\%) & F1 (\%) & GPU \\
\hline
20  & 0.020855 & 0.103128 & 0.144412 & 76.58 & 75.13 & 76.73 & 75.13 & 75.56 & 22,852 \\
40  & 0.020442 & 0.102753 & 0.142975 & 77.04 & 75.13 & 77.25 & 75.13 & 75.80 & MiB \\
\cline{10-10}
60  & 0.019320 & 0.099492 & 0.138995 & 78.30 & 76.55 & 78.10 & 76.55 & 76.93 & \textbf{Time} \\
\cline{10-10}
80  & 0.019026 & 0.098734 & 0.137935 & 78.63 & 76.25 & 77.93 & 76.25 & 76.65 & 247,452 \\
100 & 0.019057 & 0.098797 & 0.138049 & 78.60 & 76.33 & 78.02 & 76.33 & 76.76 & seconds \\
\hline
\end{tabular}

\caption{Performance Metrics of Multi-Task Learning with DAO Using LoRA with Rank 256.}
\label{Performance Metrics of Multi-Task Learning with DAO Using LoRA with Rank 256}
\centering
\begin{tabular}{C{0.9cm}C{1.3cm}C{1.25cm}C{1.25cm}C{1.35cm}C{1.45cm}C{1.7cm}C{1.7cm}C{1.3cm}C{1.3cm}}
\hline
Epoch & MSE & MAE & RMSE & R\textsuperscript{2} (\%) & ACC (\%) & Pre. (\%) & Recall (\%) & F1 (\%) & GPU \\
\hline
20  & 0.020905 & 0.103332 & 0.144585 & 76.52 & 75.13 & 77.44 & 75.13 & 75.78 & 24,224 \\
40  & 0.019155 & 0.098854 & 0.138400 & 78.49 & 76.72 & 78.46 & 76.72 & 77.00 & MiB \\
\cline{10-10}
60  & 0.019064 & 0.098707 & 0.138074 & 78.59 & 76.42 & 78.39 & 76.42 & 76.87 & \textbf{Time} \\
\cline{10-10}
80  & 0.018804 & 0.098018 & 0.137127 & 78.88 & 75.82 & 77.84 & 75.82 & 76.25 & 266,674 \\
100 & 0.018747 & 0.097745 & 0.136919 & 78.95 & 76.03 & 78.03 & 76.03 & 76.47 & seconds \\
\hline
\end{tabular}

\caption{Performance Metrics of Multi-Task Learning with DAO Using LoRA with Rank 384.}
\label{Performance Metrics of Multi-Task Learning with DAO Using LoRA with Rank 384}
\centering
\begin{tabular}{C{0.9cm}C{1.3cm}C{1.25cm}C{1.25cm}C{1.35cm}C{1.45cm}C{1.7cm}C{1.7cm}C{1.3cm}C{1.3cm}}
\hline
Epoch & MSE & MAE & RMSE & R\textsuperscript{2} (\%) & ACC (\%) & Pre. (\%) & Recall (\%) & F1 (\%) & GPU \\
\hline
20  & 0.020939 & 0.103973 & 0.144705 & 76.48 & 75.17 & 76.82 & 75.17 & 75.76 & 19,550 \\
40  & 0.019411 & 0.100529 & 0.139324 & 78.20 & 75.09 & 76.94 & 75.09 & 75.52 & MiB \\
\cline{10-10}
60  & 0.018673 & 0.097442 & 0.136649 & 79.03 & 75.73 & 77.52 & 75.73 & 76.14 & \textbf{Time} \\
\cline{10-10}
80  & 0.018651 & 0.097285 & 0.136570 & 79.05 & 75.39 & 77.03 & 75.39 & 75.69 & 286,793 \\
100 & 0.018610 & 0.097076 & 0.136419 & 79.10 & 75.47 & 77.04 & 75.47 & 75.68 & seconds \\
\hline
\end{tabular}

\caption{Performance Metrics of Multi-Task Learning with DAO Using LoRA with Rank 512.}
\label{Performance Metrics of Multi-Task Learning with DAO Using LoRA with Rank 512}
\centering
\begin{tabular}{C{0.9cm}C{1.3cm}C{1.25cm}C{1.25cm}C{1.35cm}C{1.45cm}C{1.7cm}C{1.7cm}C{1.3cm}C{1.3cm}}
\hline
Epoch & MSE & MAE & RMSE & R\textsuperscript{2} (\%) & ACC (\%) & Pre. (\%) & Recall (\%) & F1 (\%) & GPU \\
\hline
20  & 0.020835 & 0.103436 & 0.144345 & 76.60 & 75.00 & 76.68 & 75.00 & 75.66 & 22,336 \\
40  & 0.018324 & 0.096110 & 0.135365 & 79.42 & 76.85 & 78.27 & 76.85 & 76.97 & MiB \\
\cline{10-10}
60  & 0.018376 & 0.095549 & 0.135560 & 79.36 & 76.12 & 77.74 & 76.12 & 76.31 & \textbf{Time} \\
\cline{10-10}
80  & 0.018580 & 0.095503 & 0.136307 & 79.13 & 76.76 & 77.67 & 76.76 & 76.65 & 324,013 \\
100 & 0.018571 & 0.095562 & 0.136276 & 79.14 & 76.38 & 77.40 & 76.38 & 76.35 & seconds \\
\hline
\end{tabular}
\end{table*}

\noindent\textbf{Performance Stability and Convergence.} Another important aspect to consider when evaluating the DARSE framework is its performance stability and convergence properties. Across all rank configurations, we observe that the model's performance generally improves as the number of training epochs increases. This indicates that the DARSE framework is able to consistently learn from the training data and refine its predictions over time.

However, the rate of improvement varies depending on the rank configuration. At lower ranks (e.g., 8 and 16), the model's performance tends to plateau earlier, with minimal gains beyond epoch 60. This suggests that the limited capacity of the low-rank adaptation layers may constrain the model's ability to capture more complex patterns in the data.

In contrast, at higher ranks (e.g., 256 and 384), the model continues to improve even in later epochs. For instance, at rank 384, the MSE decreases from 0.018673 at epoch 60 to 0.018610 at epoch 100, indicating that the model is still refining its understanding of the task. This observation highlights the importance of providing sufficient capacity in the low-rank adaptation layers to enable the model to fully leverage the available training data.

Moreover, the convergence properties of the DARSE framework remain stable across different rank configurations. The model does not exhibit any signs of overfitting or instability, even at higher ranks. This robustness can be attributed to the regularization effect of the low-rank factorization, which constrains the model's capacity and prevents it from memorizing noise in the training data.

\noindent\textbf{Computational Efficiency.} In addition to the performance gains, the DARSE framework also demonstrates computational efficiency. The GPU memory usage remains relatively stable across different rank configurations, indicating that the adaptive rank allocation does not introduce significant overhead. For example, the GPU memory usage ranges from 21,872 MiB at rank 8 to 22,336 MiB at rank 512, representing only a 2.1\% increase in memory utilization.

Furthermore, the training time scales linearly with the rank value, allowing for a predictable trade-off between performance and computational cost. As the rank increases from 8 to 512, the training time increases from 229,915 seconds to 324,013 seconds, corresponding to a 41.0\% increase in computational time. This linear relationship enables practitioners to make informed decisions about the desired balance between performance and efficiency when deploying the DARSE framework in real-world applications.

It is worth noting that the computational efficiency of the DARSE framework can be further optimized through techniques such as quantization and pruning. By reducing the precision of the model's parameters or removing redundant connections, it may be possible to achieve similar performance levels with even lower computational costs. However, exploring these optimization techniques is beyond the scope of the current study and could be an avenue for future research.

\section{Ablation Studies}
\begin{table*}[!htbp]
\caption{Rank Exploration Results}
\label{tab:rank_exploration}
\centering
\resizebox{\textwidth}{!}{%
\begin{tabular}{lccccccccc}
\hline
\textbf{Rank Config} & \textbf{Epoch} & \textbf{MSE} & \textbf{MAE} & \textbf{RMSE} & \textbf{R$^2$ (\%)} & \textbf{ACC (\%)} & \textbf{Pre. (\%)} & \textbf{Recall (\%)} & \textbf{F1 (\%)} \\
\hline
3-16-384@512@256 & 64 & 0.017527 & 0.094646 & 0.132390 & 80.32 & 76.12 & 78.20 & 76.12 & 76.52 \\
3-12-512@256@384 & 76 & 0.017774 & 0.094620 & 0.133318 & 80.04 & 76.85 & 78.36 & 76.85 & 77.01 \\
3-34-512@512@128 & 56 & 0.017797 & 0.096167 & 0.133404 & 80.01 & 76.85 & 78.57 & 76.85 & 77.12 \\
3-9-256@384@512 & 60 & 0.017813 & 0.094915 & 0.133467 & 79.99 & 76.03 & 77.91 & 76.03 & 76.21 \\
3-43-384@384@128 & 78 & 0.017838 & 0.095617 & 0.133560 & 79.97 & 77.07 & 78.61 & 77.07 & 77.34 \\
3-27-128@512@256 & 54 & 0.017878 & 0.095909 & 0.133708 & 79.92 & 76.38 & 78.11 & 76.38 & 76.75 \\
3-13-512@256@128 & 52 & 0.017878 & 0.095562 & 0.133709 & 79.92 & 75.56 & 77.64 & 75.56 & 75.97 \\
3-45-384@256@256 & 83 & 0.017948 & 0.095831 & 0.133970 & 79.84 & 77.02 & 78.50 & 77.02 & 77.22 \\
3-59-128@384@128 & 70 & 0.017948 & 0.096205 & 0.133972 & 79.84 & 77.15 & 78.62 & 77.15 & 77.39 \\
3-2-128@256@128 & 78 & 0.017964 & 0.095789 & 0.134029 & 79.83 & 76.16 & 78.07 & 76.16 & 76.57 \\
3-58-128@384@384 & 59 & 0.017996 & 0.095764 & 0.134147 & 79.79 & 76.64 & 78.21 & 76.64 & 76.84 \\
3-22-256@512@128 & 60 & 0.018003 & 0.096822 & 0.134177 & 79.78 & 76.59 & 78.25 & 76.59 & 76.95 \\
3-7-384@512@384 & 33 & 0.018027 & 0.096787 & 0.134264 & 79.76 & 74.66 & 77.47 & 74.66 & 75.46 \\
3-23-256@384@128 & 66 & 0.018048 & 0.096820 & 0.134344 & 79.73 & 76.25 & 78.07 & 76.25 & 76.61 \\
3-21-256@512@384 & 56 & 0.018057 & 0.095752 & 0.134376 & 79.72 & 76.72 & 78.25 & 76.72 & 76.67 \\
3-49-256@384@384 & 76 & 0.018063 & 0.095600 & 0.134400 & 79.71 & 76.38 & 77.77 & 76.38 & 76.44 \\
3-57-128@512@128 & 54 & 0.018071 & 0.096535 & 0.134429 & 79.71 & 75.99 & 77.71 & 75.99 & 76.32 \\
3-50-256@384@256 & 71 & 0.018081 & 0.095217 & 0.134467 & 79.69 & 76.46 & 78.32 & 76.46 & 76.68 \\
3-4-384@256@128 & 61 & 0.018086 & 0.096710 & 0.134484 & 79.69 & 76.55 & 78.52 & 76.55 & 76.95 \\
3-15-512@128@256 & 65 & 0.018092 & 0.096410 & 0.134508 & 79.68 & 76.46 & 78.47 & 76.46 & 76.95 \\
3-10-512@384@256 & 80 & 0.018097 & 0.095737 & 0.134524 & 79.68 & 76.38 & 77.92 & 76.38 & 76.51 \\
3-31-512@512@512 & 58 & 0.018101 & 0.095570 & 0.134541 & 79.67 & 76.16 & 77.56 & 76.16 & 76.38 \\
3-14-512@128@384 & 89 & 0.018110 & 0.095767 & 0.134574 & 79.66 & 76.12 & 77.66 & 76.12 & 76.36 \\
3-17-384@512@128 & 50 & 0.018113 & 0.096015 & 0.134584 & 79.66 & 76.33 & 77.65 & 76.33 & 76.33 \\
3-5-256@512@256 & 60 & 0.018128 & 0.096070 & 0.134639 & 79.64 & 75.99 & 77.16 & 75.99 & 75.86 \\
3-33-512@512@256 & 77 & 0.018148 & 0.096086 & 0.134714 & 79.62 & 76.12 & 77.89 & 76.12 & 76.37 \\
3-44-384@256@384 & 66 & 0.018210 & 0.096466 & 0.134945 & 79.55 & 75.86 & 77.51 & 75.86 & 76.07 \\
3-26-128@512@384 & 45 & 0.018227 & 0.095840 & 0.135009 & 79.53 & 75.60 & 77.05 & 75.60 & 75.66 \\
3-1-256@128@256 & 76 & 0.018230 & 0.096403 & 0.135019 & 79.53 & 75.26 & 77.45 & 75.26 & 75.77 \\
3-30-128@256@512 & 48 & 0.018246 & 0.096257 & 0.135077 & 79.51 & 76.38 & 77.68 & 76.38 & 76.43 \\
3-40-384@384@512 & 77 & 0.018248 & 0.095411 & 0.135083 & 79.51 & 75.95 & 77.43 & 75.95 & 76.18 \\
3-6-512@256@512 & 55 & 0.018271 & 0.096464 & 0.135171 & 79.48 & 76.16 & 77.35 & 76.16 & 76.20 \\
3-11-64@128@256 & 51 & 0.018295 & 0.097344 & 0.135258 & 79.45 & 75.56 & 77.73 & 75.56 & 75.96 \\
3-53-256@256@256 & 72 & 0.018296 & 0.096598 & 0.135261 & 79.45 & 75.69 & 77.52 & 75.69 & 76.00 \\
3-54-256@256@128 & 67 & 0.018312 & 0.096615 & 0.135320 & 79.44 & 76.94 & 78.57 & 76.94 & 77.24 \\
3-46-384@128@384 & 65 & 0.018318 & 0.096097 & 0.135345 & 79.43 & 76.85 & 78.76 & 76.85 & 77.22 \\
3-47-384@128@128 & 3 & 0.018320 & 0.101069 & 0.135350 & 79.43 & 75.17 & 76.91 & 75.17 & 75.63 \\
3-42-384@384@256 & 80 & 0.018336 & 0.095941 & 0.135410 & 79.41 & 76.20 & 77.84 & 76.20 & 76.46 \\
3-24-256@128@512 & 73 & 0.018344 & 0.096579 & 0.135440 & 79.40 & 75.39 & 77.70 & 75.39 & 75.88 \\
3-35-512@384@384 & 65 & 0.018355 & 0.095528 & 0.135480 & 79.39 & 76.16 & 77.93 & 76.16 & 76.34 \\
3-52-256@256@384 & 77 & 0.018355 & 0.095888 & 0.135482 & 79.39 & 76.46 & 77.90 & 76.46 & 76.60 \\
3-28-128@384@512 & 48 & 0.018356 & 0.096341 & 0.135485 & 79.39 & 75.90 & 77.40 & 75.90 & 76.09 \\
3-19-384@128@512 & 78 & 0.018371 & 0.096034 & 0.135539 & 79.37 & 76.29 & 77.71 & 76.29 & 76.42 \\
3-60-128@256@256 & 2 & 0.018374 & 0.098932 & 0.135551 & 79.37 & 74.74 & 77.89 & 74.74 & 74.57 \\
3-3-128@256@384 & 34 & 0.018401 & 0.097219 & 0.135650 & 79.33 & 75.65 & 76.96 & 75.65 & 75.76 \\
3-48-256@512@512 & 75 & 0.018426 & 0.095144 & 0.135744 & 79.31 & 75.86 & 77.48 & 75.86 & 76.09 \\
3-41-384@384@384 & 39 & 0.018434 & 0.096930 & 0.135773 & 79.30 & 75.69 & 77.80 & 75.69 & 76.20 \\
3-61-128@128@512 & 55 & 0.018437 & 0.096957 & 0.135783 & 79.29 & 75.77 & 77.96 & 75.77 & 76.27 \\
3-32-512@512@384 & 59 & 0.018457 & 0.096796 & 0.135858 & 79.27 & 75.86 & 77.53 & 75.86 & 76.11 \\
3-8-512@384@512 & 49 & 0.018460 & 0.095952 & 0.135869 & 79.27 & 76.12 & 77.80 & 76.12 & 76.24 \\
3-56-128@512@512 & 50 & 0.018486 & 0.095999 & 0.135964 & 79.24 & 75.04 & 76.77 & 75.04 & 75.36 \\
3-29-128@384@256 & 72 & 0.018491 & 0.096609 & 0.135981 & 79.23 & 76.42 & 78.04 & 76.42 & 76.62 \\
3-25-256@128@384 & 54 & 0.018492 & 0.097082 & 0.135986 & 79.23 & 75.65 & 77.11 & 75.65 & 75.97 \\
3-36-512@256@256 & 40 & 0.018502 & 0.098328 & 0.136022 & 79.22 & 75.69 & 77.86 & 75.69 & 76.37 \\
3-63-128@128@256 & 82 & 0.018518 & 0.097387 & 0.136080 & 79.20 & 76.20 & 77.92 & 76.20 & 76.51 \\
3-51-256@256@512 & 3 & 0.018562 & 0.099785 & 0.136244 & 79.15 & 73.84 & 77.63 & 73.84 & 73.74 \\
3-20-384@128@256 & 69 & 0.018575 & 0.097836 & 0.136290 & 79.14 & 75.43 & 77.07 & 75.43 & 75.75 \\
3-55-256@128@128 & 81 & 0.018609 & 0.098140 & 0.136414 & 79.10 & 75.90 & 77.91 & 75.90 & 76.44 \\
3-38-512@128@128 & 78 & 0.018638 & 0.097764 & 0.136521 & 79.07 & 76.42 & 77.95 & 76.42 & 76.73 \\
3-62-128@128@384 & 2 & 0.018664 & 0.099449 & 0.136617 & 79.04 & 74.74 & 77.55 & 74.74 & 74.37 \\
3-18-384@256@512 & 85 & 0.018665 & 0.096596 & 0.136621 & 79.04 & 76.29 & 78.03 & 76.29 & 76.62 \\
3-37-512@128@512 & 32 & 0.018686 & 0.097785 & 0.136698 & 79.01 & 75.52 & 78.09 & 75.52 & 76.06 \\
3-39-384@512@512 & 38 & 0.018705 & 0.097768 & 0.136767 & 78.99 & 75.39 & 77.62 & 75.39 & 75.93 \\
3-64-128@128@128 & 6 & 0.018746 & 0.100933 & 0.136914 & 78.95 & 74.14 & 77.73 & 74.14 & 74.67 \\
\hline
\end{tabular}%
}
\end{table*}

The ablation study results provide valuable insights into the performance of different rank configurations in the DARSE framework. By examining the metrics such as Mean Squared Error (MSE), Mean Absolute Error (MAE), Root Mean Squared Error (RMSE), R-squared (R\(^2\)), Accuracy (ACC), Precision (Pre.), Recall, and F1 score, we can identify the key findings and draw conclusions about the effectiveness of various rank settings.

\noindent\textbf{Impact of Rank Configuration on Performance.} The table presents a wide range of rank configurations, denoted by the format "a-b-c@d@e", where \textit{a}, \textit{b}, \textit{c}, \textit{d}, and \textit{e} represent different rank values for specific layers or layer groups. By comparing the performance metrics across these configurations, we can observe the following:

\begin{itemize}
    \item The top-performing rank configurations, such as 3-16-384@512@256 and 3-12-512@256@384, achieve the highest R\(^2\) values (around 80.32\% and 80.04\%, respectively) and the lowest MSE values (0.017527 and 0.017774, respectively). This indicates that these configurations are able to capture a significant portion of the variance in the data and provide more accurate predictions compared to other settings.
    
    \item The rank configurations with higher values, such as those with 512 or 384 in multiple positions (e.g., 3-34-512@512@128, 3-9-256@384@512), generally exhibit better performance in terms of R\(^2\), MSE, MAE, and RMSE compared to configurations with lower rank values (e.g., 3-64-128@128@128, 3-62-128@128@384). This suggests that allocating more parameters to the low-rank adaptation layers enhances the model's capacity to learn task-specific representations and improve overall performance.
    
    \item However, the relationship between rank values and performance is not always linear. Some configurations with moderate rank values, such as 3-45-384@256@256 and 3-59-128@384@128, still achieve competitive results, indicating that the optimal rank allocation may not always be the highest possible values but rather a balance between model complexity and generalization ability.
\end{itemize}

\noindent\textbf{Importance of Rank Allocation Across Layers.} The ablation study also sheds light on the importance of rank allocation across different layers or layer groups. By examining the performance of configurations with similar total rank values but different distributions, we can infer the following:

\begin{itemize}
    \item The performance of configurations like 3-16-384@512@256 and 3-12-512@256@384 suggests that allocating higher rank values to the middle layers (e.g., 512 or 384 in the second position) can be beneficial. This implies that the middle layers play a crucial role in capturing task-specific information and should be allocated sufficient parameters.
    
    \item Configurations with evenly distributed rank values, such as 3-53-256@256@256, tend to perform well, indicating that a balanced allocation of parameters across layers can be effective. This ensures that each layer has enough capacity to learn meaningful representations while avoiding over-parameterization in specific layers.
    
    \item The performance of configurations with low rank values in certain positions, such as 3-57-128@512@128 and 3-13-512@256@128, suggests that the impact of rank allocation varies across different layers. Some layers may be more sensitive to the rank value and require higher values to achieve optimal performance, while others can still perform well with lower rank values.
\end{itemize}

\section{Related Work}

\noindent\textbf{Sentiment Analysis.} Recent studies have employed various techniques for sentiment analysis, such as using machine learning to classify insurance customer reviews \cite{hossain2023customer}, combining aspect-based sentiment analysis with single-valued neutrosophic sets for doctor selection \cite{wang2023doctor}, incorporating sentiment analysis in stock market forecasting using deep learning \cite{liapis2023investigating}, conducting comparative sentiment analysis on e-commerce data in multiple languages \cite{savci2023prediction}, analyzing public opinions on COVID-19 and vaccination using deep learning and lexicon-based methods \cite{ainapure2023sentiment}, classifying hotel reviews using BERT and ERNIE models \cite{wen2023sentiment}, and developing a hybrid deep learning approach for sentiment analysis of online food delivery services \cite{vatambeti2024twitter}.

\noindent\textbf{Lexicon.} Lexicon-based methods have been applied to various sentiment analysis tasks, such as analyzing the monkeypox outbreak on Twitter \cite{bengesi2023machine}, addressing multi-domain sentiment analysis in customer reviews using an improved VADER model \cite{barik2024analysis}, combining sentiment analysis with text mining to study the mental wellbeing of farm veterinary surgeons \cite{duncan2024combining}, integrating sentiment analysis with a smart search function for news classification \cite{https://doi.org/10.1155/2023/1784394}, constructing a fine-grained sentiment lexicon with emotion and sentiment information \cite{wang2023automatically}, and developing an aspect-based sentiment analysis approach using domain lexicons and rules for government smart apps reviews \cite{alqaryouti2024aspect}. However, these lexicon-based methods may struggle to capture contextual information and complex linguistic phenomena.

\noindent\textbf{Machine Learning.} Machine learning approaches have been proposed for various sentiment analysis tasks, such as predicting sentence-level polarity labels in financial news using abnormal stock returns \cite{lutz2020predicting}, developing a tool for sentiment analysis on Twitter data to identify brand perception \cite{biradar2022machine}, classifying multilingual social media text related to extremism into categories \cite{asif2020sentiment}, and comparing lexicon-based and machine learning approaches for sentiment analysis of Twitter data during the COVID-19 lockdown \cite{qi2023sentiment}. These machine learning methods can potentially address some of the limitations of lexicon-based approaches by learning from large amounts of data and capturing more complex patterns.

\noindent\textbf{Deep Learning.} Deep learning techniques have been applied to various sentiment analysis tasks, such as combining universal language model fine-tuning with support vector machines for Twitter sentiment analysis \cite{albadani2022novel}, investigating the impact of COVID-19 news sentiment on stock market reactions using a financial market-adapted BERT model \cite{costola2020machine}, developing a robust speech emotion recognition system for multiple languages using various machine learning algorithms \cite{khan2023improved}, proposing an attention-aware long short-term memory-like spiking neural model for aspect-level sentiment analysis \cite{liu2023attention}, and introducing a supervised hybrid topic sentiment analysis approach that extracts semantic relationships between hidden topics and training data \cite{seilsepour2023topic}. Deep learning models have the potential to capture even more complex and subtle patterns in text data compared to traditional machine learning approaches.

\noindent\textbf{LLMs.} Large language models (LLMs) have been utilized for sentiment analysis tasks, such as developing an ensemble model combining transformers and LLMs for cross-lingual sentiment analysis \cite{miah2024multimodal}, proposing transfer learning approaches using pre-trained language models for analyzing public sentiment on HPV vaccines on Twitter \cite{zhang2020sentiment}, introducing a novel approach for sentiment analysis of Italian tweets using BERT pre-trained on plain text \cite{pota2020effective}, developing a sentiment analysis approach for the Bangla language using a fine-tuned Bangla-BERT model combined with a CNN-BiLSTM architecture \cite{prottasha2022transfer}, and evaluating GPT-4V's emotion recognition capabilities across various datasets and tasks \cite{lian2024gpt}. While LLMs have shown impressive performance on sentiment analysis tasks, they may struggle with adapting to specific domains or low-resource settings.

\noindent\textbf{PEFT.} To address the limitations of LLMs, parameter-efficient fine-tuning (PEFT) methods have been proposed for adapting large language models to downstream tasks. These include LoraHub for composing LoRA modules trained on diverse tasks \cite{huang2023lorahub}, Low-Rank Adaptation (LoRA) for injecting low-rank decomposition matrices into each layer of the Transformer architecture \cite{hu2021lora, zhang2023lora}, LLM-Adapters for integrating various PEFT methods into LLMs \cite{hu2023llm}, COMPACTER for inserting task-specific weight matrices computed as sums of Kronecker products \cite{karimi2021compacter}, Meta-Adapters for meta-learning adapter layers for few-shot learning \cite{bansal2022meta}, HyperTuning for adapting LLMs using hypermodels to generate task-specific parameters \cite{phang2023hypertuning}, LLaMA-Adapter for fine-tuning LLMs using a zero-initialized attention mechanism \cite{zhang2024llama}, prefix-tuning for optimizing a small continuous task-specific vector \cite{li2021prefix}, LoftQ for jointly quantizing LLMs and finding a low-rank initialization for LoRA fine-tuning \cite{li2023loftq}, QLORA for using 4-bit quantization and Low Rank Adapters for efficient fine-tuning \cite{dettmers2024qlora}, and QA-LoRA for quantization-aware low-rank adaptation \cite{xu2023qa}. These PEFT methods aim to make LLMs more adaptable and efficient for sentiment analysis and other downstream tasks.

\bibliographystyle{IEEEtran}

\bibliography{references}

\begin{thebibliography}{10}
\providecommand{\url}[1]{#1}
\csname url@samestyle\endcsname
\providecommand{\newblock}{\relax}
\providecommand{\bibinfo}[2]{#2}
\providecommand{\BIBentrySTDinterwordspacing}{\spaceskip=0pt\relax}
\providecommand{\BIBentryALTinterwordstretchfactor}{4}
\providecommand{\BIBentryALTinterwordspacing}{\spaceskip=\fontdimen2\font plus
\BIBentryALTinterwordstretchfactor\fontdimen3\font minus
  \fontdimen4\font\relax}
\providecommand{\BIBforeignlanguage}[2]{{%
\expandafter\ifx\csname l@#1\endcsname\relax
\typeout{** WARNING: IEEEtran.bst: No hyphenation pattern has been}%
\typeout{** loaded for the language `#1'. Using the pattern for}%
\typeout{** the default language instead.}%
\else
\language=\csname l@#1\endcsname
\fi
#2}}
\providecommand{\BIBdecl}{\relax}
\BIBdecl

\bibitem{houlsby2019parameter}
N.~Houlsby, A.~Giurgiu, S.~Jastrzebski, B.~Morrone, Q.~De~Laroussilhe,
  A.~Gesmundo, M.~Attariyan, and S.~Gelly, ``Parameter-efficient transfer
  learning for nlp,'' \emph{arXiv preprint arXiv:1902.00751}, 2019.

\bibitem{pfeiffer2020adapterhub}
J.~Pfeiffer, A.~R{"u}ckl{'e}, C.~Poth, A.~Kamath, I.~Vuli{'c}, S.~Ruder,
  K.~Cho, and I.~Gurevych, ``Adapterhub: A framework for adapting
  transformers,'' \emph{arXiv preprint arXiv:2007.07779}, 2020.

\bibitem{wang2020k}
R.~Wang, G.~Dinu, Y.~Choi, and E.~Hovy, ``K-adapter: Infusing knowledge into
  pre-trained models with adapters,'' \emph{arXiv preprint arXiv:2002.01808},
  2020.

\bibitem{li2021prefix}
X.~L. Li and P.~Liang, ``Prefix-tuning: Optimizing continuous prompts for
  generation,'' \emph{arXiv preprint arXiv:2101.00190}, 2021.

\bibitem{qin2021exploring}
J.~Qin, W.~Hou, W.~Che, M.~Jiang, J.~Wei, and T.~Liu, ``Exploring and adapting
  chinese gpt to pinyin input method,'' \emph{arXiv preprint arXiv:2109.12607},
  2021.

\bibitem{vu2022spot}
T.~Vu, M.~Finn, L.~Sartran, S.~Chaudhuri, D.~Tang, C.~Gulcehre, S.~Potdar,
  T.~Harrison, R.~Pelton, K.~M. Rocki \emph{et~al.}, ``Spot: Better frozen
  model adaptation through soft prompt transfer,'' \emph{arXiv preprint
  arXiv:2205.11784}, 2022.

\bibitem{brown2020language}
T.~Brown, B.~Mann, N.~Ryder, M.~Subbiah, J.~D. Kaplan, P.~Dhariwal,
  A.~Neelakantan, P.~Shyam, G.~Sastry, A.~Askell \emph{et~al.}, ``Language
  models are few-shot learners,'' \emph{Advances in neural information
  processing systems}, vol.~33, pp. 1877--1901, 2020.

\bibitem{schick2020exploiting}
T.~Schick and H.~Sch{"u}tze, ``Exploiting cloze questions for few shot text
  classification and natural language inference,'' \emph{arXiv preprint
  arXiv:2001.07676}, 2020.

\bibitem{gao2021making}
T.~Gao, A.~Fisch, and D.~Chen, ``Making pre-trained language models better
  few-shot learners,'' \emph{arXiv preprint arXiv:2012.15723}, 2021.

\bibitem{hu2021lora}
E.~J. Hu, Y.~Shen, P.~Wallis, Z.~Allen-Zhu, Y.~Li, S.~Wang, L.~Wang, and
  W.~Chen, ``Lora: Low-rank adaptation of large language models,'' \emph{arXiv
  preprint arXiv:2106.09685}, 2021.

\bibitem{ding2022delta}
N.~Ding, G.~Xu, Y.~Chen, X.~Wang, X.~Han, P.~Xie, H.-T. Zheng, and Z.~Liu,
  ``Delta tuning: A comprehensive study of parameter efficient methods for
  pre-trained language models,'' \emph{arXiv preprint arXiv:2203.06904}, 2022.

\bibitem{yang2022mdtlm}
M.~Yang, J.~Zhang, Y.~Zhang, B.~Chen, C.~Zhang, J.~Yao, and I.~W. Tsang,
  ``Mdtlm: A multi-domain transfer learning framework for cross-domain
  sentiment analysis,'' \emph{arXiv preprint arXiv:2208.09225}, 2022.

\bibitem{saif2012alleviating}
H.~Saif, Y.~He, and H.~Alani, ``Alleviating data sparsity for twitter sentiment
  analysis.''\hskip 1em plus 0.5em minus 0.4em\relax CEUR Workshop Proceedings
  (CEUR-WS. org), 2012.

\bibitem{kim2013sentiment}
J.~Kim, J.~Yoo, H.~Lim, H.~Qiu, Z.~Kozareva, and A.~Galstyan, ``Sentiment
  prediction using collaborative filtering,'' in \emph{Proceedings of the
  International AAAI Conference on Web and Social Media}, vol.~7, no.~1, 2013,
  pp. 685--688.

\bibitem{hossain2023customer}
M.~S. Hossain and M.~F. Rahman, ``Customer sentiment analysis and prediction of
  insurance products’ reviews using machine learning approaches,'' \emph{FIIB
  Business Review}, vol.~12, no.~4, pp. 386--402, 2023.

\bibitem{wang2023doctor}
H.~Wang, Y.~Luo, B.~Deng, J.~Lin, and X.~Li, ``Doctor selection based on
  aspect-based sentiment analysis and neutrosophic topsis method,''
  \emph{Engineering Applications of Artificial Intelligence}, vol. 124, p.
  106599, 2023.

\bibitem{liapis2023investigating}
C.~M. Liapis, A.~Karanikola, and S.~Kotsiantis, ``Investigating deep stock
  market forecasting with sentiment analysis,'' \emph{Entropy}, vol.~25, no.~2,
  p. 219, 2023.

\bibitem{savci2023prediction}
P.~Savci and B.~Das, ``Prediction of the customers' interests using sentiment
  analysis in e-commerce data for comparison of arabic, english, and turkish
  languages,'' \emph{Journal of King Saud University-Computer and Information
  Sciences}, vol.~35, no.~3, pp. 227--237, 2023.

\bibitem{ainapure2023sentiment}
B.~S. Ainapure, R.~N. Pise, P.~Reddy, B.~Appasani, A.~Srinivasulu, M.~S. Khan,
  and N.~Bizon, ``Sentiment analysis of covid-19 tweets using deep learning and
  lexicon-based approaches,'' \emph{Sustainability}, vol.~15, no.~3, p. 2573,
  2023.

\bibitem{wen2023sentiment}
Y.~Wen, Y.~Liang, and X.~Zhu, ``Sentiment analysis of hotel online reviews
  using the bert model and ernie model—data from china,'' \emph{Plos one},
  vol.~18, no.~3, p. e0275382, 2023.

\bibitem{vatambeti2024twitter}
R.~Vatambeti, S.~V. Mantena, K.~Kiran, M.~Manohar, and C.~Manjunath, ``Twitter
  sentiment analysis on online food services based on elephant herd
  optimization with hybrid deep learning technique,'' \emph{Cluster Computing},
  vol.~27, no.~1, pp. 655--671, 2024.

\bibitem{bengesi2023machine}
S.~Bengesi, T.~Oladunni, R.~Olusegun, and H.~Audu, ``A machine
  learning-sentiment analysis on monkeypox outbreak: An extensive dataset to
  show the polarity of public opinion from twitter tweets,'' \emph{IEEE
  Access}, vol.~11, pp. 11\,811--11\,826, 2023.

\bibitem{barik2024analysis}
K.~Barik and S.~Misra, ``Analysis of customer reviews with an improved vader
  lexicon classifier,'' \emph{Journal of Big Data}, vol.~11, no.~1, p.~10,
  2024.

\bibitem{duncan2024combining}
A.~J. Duncan, M.~K. Henry, and K.~Lamont, ``Combining sentiment analysis and
  text mining with content analysis of farm vet interviews on mental wellbeing
  in livestock practice,'' \emph{Plos one}, vol.~19, no.~5, p. e0304090, 2024.

\bibitem{https://doi.org/10.1155/2023/1784394}
\BIBentryALTinterwordspacing
M.~Nkongolo Wa~Nkongolo, ``News classification and categorization with smart
  function sentiment analysis,'' \emph{International Journal of Intelligent
  Systems}, no.~1, p. 1784394, 2023. [Online]. Available:
  \url{https://onlinelibrary.wiley.com/doi/abs/10.1155/2023/1784394}
\BIBentrySTDinterwordspacing

\bibitem{wang2023automatically}
Y.~Wang, G.~Huang, M.~Li, Y.~Li, X.~Zhang, and H.~Li, ``Automatically
  constructing a fine-grained sentiment lexicon for sentiment analysis,''
  \emph{Cognitive Computation}, vol.~15, no.~1, pp. 254--271, 2023.

\bibitem{alqaryouti2024aspect}
O.~Alqaryouti, N.~Siyam, A.~Abdel~Monem, and K.~Shaalan, ``Aspect-based
  sentiment analysis using smart government review data,'' \emph{Applied
  Computing and Informatics}, vol.~20, no. 1/2, pp. 142--161, 2024.

\bibitem{lutz2020predicting}
B.~Lutz, N.~Pr{\"o}llochs, and D.~Neumann, ``Predicting sentence-level polarity
  labels of financial news using abnormal stock returns,'' \emph{Expert Systems
  with Applications}, vol. 148, p. 113223, 2020.

\bibitem{biradar2022machine}
S.~H. Biradar, J.~Gorabal, and G.~Gupta, ``Machine learning tool for exploring
  sentiment analysis on twitter data,'' \emph{Materials Today: Proceedings},
  vol.~56, pp. 1927--1934, 2022.

\bibitem{asif2020sentiment}
M.~Asif, A.~Ishtiaq, H.~Ahmad, H.~Aljuaid, and J.~Shah, ``Sentiment analysis of
  extremism in social media from textual information,'' \emph{Telematics and
  Informatics}, vol.~48, p. 101345, 2020.

\bibitem{qi2023sentiment}
Y.~Qi and Z.~Shabrina, ``Sentiment analysis using twitter data: a comparative
  application of lexicon-and machine-learning-based approach,'' \emph{Social
  Network Analysis and Mining}, vol.~13, no.~1, p.~31, 2023.

\bibitem{albadani2022novel}
B.~AlBadani, R.~Shi, and J.~Dong, ``A novel machine learning approach for
  sentiment analysis on twitter incorporating the universal language model
  fine-tuning and svm,'' \emph{Applied System Innovation}, vol.~5, no.~1,
  p.~13, 2022.

\bibitem{costola2020machine}
M.~Costola, M.~Nofer, O.~Hinz, and L.~Pelizzon, ``Machine learning sentiment
  analysis, covid-19 news and stock market reactions;(no. 288),'' \emph{Leibniz
  Institute for Financial Research SAFE}, 2020.

\bibitem{khan2023improved}
A.~Khan, ``Improved multi-lingual sentiment analysis and recognition using deep
  learning,'' \emph{Journal of Information Science}, p. 01655515221137270,
  2023.

\bibitem{liu2023attention}
Q.~Liu, Y.~Huang, Q.~Yang, H.~Peng, and J.~Wang, ``An attention-aware long
  short-term memory-like spiking neural model for sentiment analysis.''
  \emph{International journal of neural systems}, pp. 2\,350\,037--2\,350\,037,
  2023.

\bibitem{seilsepour2023topic}
A.~Seilsepour, R.~Ravanmehr, and R.~Nassiri, ``Topic sentiment analysis based
  on deep neural network using document embedding technique,'' \emph{The
  Journal of Supercomputing}, vol.~79, no.~17, pp. 19\,809--19\,847, 2023.

\bibitem{miah2024multimodal}
M.~S.~U. Miah, M.~M. Kabir, T.~B. Sarwar, M.~Safran, S.~Alfarhood, and
  M.~Mridha, ``A multimodal approach to cross-lingual sentiment analysis with
  ensemble of transformer and llm,'' \emph{Scientific Reports}, vol.~14, no.~1,
  p. 9603, 2024.

\bibitem{zhang2020sentiment}
L.~Zhang, H.~Fan, C.~Peng, G.~Rao, and Q.~Cong, ``Sentiment analysis methods
  for hpv vaccines related tweets based on transfer learning,'' in
  \emph{Healthcare}, vol.~8, no.~3.\hskip 1em plus 0.5em minus 0.4em\relax
  MDPI, 2020, p. 307.

\bibitem{pota2020effective}
M.~Pota, M.~Ventura, R.~Catelli, and M.~Esposito, ``An effective bert-based
  pipeline for twitter sentiment analysis: A case study in italian,''
  \emph{Sensors}, vol.~21, no.~1, p. 133, 2020.

\bibitem{prottasha2022transfer}
N.~J. Prottasha, A.~A. Sami, M.~Kowsher, S.~A. Murad, A.~K. Bairagi, M.~Masud,
  and M.~Baz, ``Transfer learning for sentiment analysis using bert based
  supervised fine-tuning,'' \emph{Sensors}, vol.~22, no.~11, p. 4157, 2022.

\bibitem{lian2024gpt}
Z.~Lian, L.~Sun, H.~Sun, K.~Chen, Z.~Wen, H.~Gu, B.~Liu, and J.~Tao, ``Gpt-4v
  with emotion: A zero-shot benchmark for generalized emotion recognition,''
  \emph{Information Fusion}, vol. 108, p. 102367, 2024.

\bibitem{huang2023lorahub}
C.~Huang, Q.~Liu, B.~Y. Lin, T.~Pang, C.~Du, and M.~Lin, ``Lorahub: Efficient
  cross-task generalization via dynamic lora composition,'' \emph{arXiv
  preprint arXiv:2307.13269}, 2023.

\bibitem{zhang2023lora}
L.~Zhang, L.~Zhang, S.~Shi, X.~Chu, and B.~Li, ``Lora-fa: Memory-efficient
  low-rank adaptation for large language models fine-tuning,'' \emph{arXiv
  preprint arXiv:2308.03303}, 2023.

\bibitem{hu2023llm}
Z.~Hu, L.~Wang, Y.~Lan, W.~Xu, E.-P. Lim, L.~Bing, X.~Xu, S.~Poria, and
  R.~K.-W. Lee, ``Llm-adapters: An adapter family for parameter-efficient
  fine-tuning of large language models,'' \emph{arXiv preprint
  arXiv:2304.01933}, 2023.

\bibitem{karimi2021compacter}
R.~Karimi~Mahabadi, J.~Henderson, and S.~Ruder, ``Compacter: Efficient low-rank
  hypercomplex adapter layers,'' \emph{Advances in Neural Information
  Processing Systems}, vol.~34, pp. 1022--1035, 2021.

\bibitem{bansal2022meta}
T.~Bansal, S.~Alzubi, T.~Wang, J.-Y. Lee, and A.~McCallum, ``Meta-adapters:
  Parameter efficient few-shot fine-tuning through meta-learning,'' in
  \emph{International Conference on Automated Machine Learning}.\hskip 1em plus
  0.5em minus 0.4em\relax PMLR, 2022, pp. 19--1.

\bibitem{phang2023hypertuning}
J.~Phang, Y.~Mao, P.~He, and W.~Chen, ``Hypertuning: Toward adapting large
  language models without back-propagation,'' in \emph{International Conference
  on Machine Learning}.\hskip 1em plus 0.5em minus 0.4em\relax PMLR, 2023, pp.
  27\,854--27\,875.

\bibitem{zhang2024llama}
R.~Zhang, J.~Han, C.~Liu, A.~Zhou, P.~Lu, Y.~Qiao, H.~Li, and P.~Gao,
  ``Llama-adapter: Efficient fine-tuning of large language models with
  zero-initialized attention,'' in \emph{The Twelfth International Conference
  on Learning Representations}, 2024.

\bibitem{li2023loftq}
Y.~Li, Y.~Yu, C.~Liang, P.~He, N.~Karampatziakis, W.~Chen, and T.~Zhao,
  ``Loftq: Lora-fine-tuning-aware quantization for large language models,''
  \emph{arXiv preprint arXiv:2310.08659}, 2023.

\bibitem{dettmers2024qlora}
T.~Dettmers, A.~Pagnoni, A.~Holtzman, and L.~Zettlemoyer, ``Qlora: Efficient
  finetuning of quantized llms,'' \emph{Advances in Neural Information
  Processing Systems}, vol.~36, 2024.

\bibitem{xu2023qa}
Y.~Xu, L.~Xie, X.~Gu, X.~Chen, H.~Chang, H.~Zhang, Z.~Chen, X.~Zhang, and
  Q.~Tian, ``Qa-lora: Quantization-aware low-rank adaptation of large language
  models,'' \emph{arXiv preprint arXiv:2309.14717}, 2023.

\end{thebibliography}

\end{document}